\documentclass[runningheads]{llncs}

\usepackage[colorlinks=true,linkcolor=blue, citecolor=blue, urlcolor=blue]{hyperref}
\usepackage[misc]{ifsym}
\usepackage[utf8]{inputenc} 
\usepackage[T1]{fontenc}    
\usepackage{url}            
\usepackage{booktabs}       
\usepackage{amsfonts}       
\usepackage{nicefrac}       
\usepackage{microtype}      
\usepackage{lipsum}		
\usepackage{graphicx}
\usepackage[square, sort, numbers,compress,sectionbib]{natbib}
\usepackage{doi}
\usepackage[table,xcdraw]{xcolor}
\usepackage{algpseudocode}
\usepackage{algorithm}
\usepackage{amsmath}
\usepackage{caption}
\usepackage{multirow}

\usepackage{colortbl}


\usepackage{caption}

\usepackage{textcomp}
\usepackage{multicol}
\usepackage{indentfirst} 
\def\ie{\textit{i.e., }}
\def\eg{\textit{e.g., }}

\usepackage
[        
        left=2.25cm,
        right=2.25cm,
        top=2cm,
        bottom=2cm,
]
{geometry}
\begin{document}
\title{Providing Insights for Open-Response Surveys \\
via End-to-End Context-Aware Clustering}
%
%


\author{Soheil Esmaeilzadeh\thanks{\Letter\,\,Corresponding Author - Published in AIED 2022, Springer vol 13355, pp 526–532, \href{https://doi.org/10.1007/978-3-031-11644-5_44}{doi:10.1007/978-3-031-11644-5}} \and
Brian Williams \and
Davood Shamsi \and
Onar Vikingstad}
\authorrunning{S. Esmaeilzadeh et al.}
%
\institute{Apple, \\ 
Cupertino, CA, USA \\
\href{mailto:soes@alumni.stanford.edu}{$^*$soes@alumni.stanford.edu}, \href{mailto:brian_d_williams@apple.com}{brian$\_$d$\_$williams@apple.com}, \\
\href{mailto:davood@apple.com}{davood@apple.com}, \href{mailto:vikingstad@apple.com}{vikingstad@apple.com}}
\maketitle              
\begin{abstract}
Teachers often conduct surveys in order to collect data from a predefined group of students to gain insights into topics of interest. When analyzing surveys with open-ended textual responses, it is extremely time-consuming, labor-intensive, and difficult to manually process all the responses into an insightful and comprehensive report. In the analysis step, traditionally, the teacher has to read each of the responses and decide on how to group them in order to extract insightful information. Even though it is possible to group the responses only using certain keywords, such an approach would be limited since it not only fails to account for embedded contexts but also cannot detect polysemous words or phrases and semantics that are not expressible in single words. In this work, we present a novel end-to-end context-aware framework that extracts, aggregates, and abbreviates embedded semantic patterns in open-response survey data. Our framework relies on a pre-trained natural language model in order to encode the textual data into semantic vectors. The encoded vectors then get clustered either into an optimally tuned number of groups or into a set of groups with pre-specified titles. In the former case, the clusters are then further analyzed to extract a representative set of keywords or summary sentences that serve as the labels of the clusters. In our framework, for the designated clusters, we finally provide context-aware wordclouds that demonstrate the semantically prominent keywords within each group. Honoring user privacy, we have successfully built the on-device implementation of our framework suitable for real-time analysis on mobile devices and have tested it on a synthetic dataset. Our framework reduces the costs at-scale by automating the process of extracting the most insightful information pieces from survey data.

\end{abstract}

\section{Introduction}
\noindent Formative assessment commonly refers to a set of activities undertaken by teachers to gather information about the learning progress of students \cite{Black1998}. Such information can then be used by teachers as feedback in order to better plan teaching and learning activities in the classroom. The main benefit of formative assessment is that its consistent use has been shown to increase the achievements of students \cite{Duckor2014}. Formative assessment can uncover the concepts that the students do not understand during the learning process. Although formative assessments can be graded, evaluations of these assessments usually do not affect the final grades of the students because their main focus is on assessing teaching effectiveness and understanding of students about the class materials. \\
\indent Surveys (slip-tickets) are a commonly used formative assessment method in classrooms \cite{Cornelius2014}. Through a survey the teacher shares a set of questions with all the students at the end of the class with a focus on the most important concepts taught during the lecture. Students have to respond to the questions before they leave the classroom. Surveys provide firsthand information about what students have understood from the lesson and whether the course objectives have been achieved. The teachers then review the responses and adjust the teaching plans for the upcoming days \cite{Dixson2016}. \\
\indent Formative assessment using surveys commonly includes four steps, namely, (i) creation, (ii) collection, (iii) analysis, and (iv) action. In the \textit{creation} step, teachers simply put together a set of questions to evaluate specific items covered during the lecture. For instance, teachers can create the surveys to check the understanding of students by asking them to summarize the key points from the lesson; or to verify that students can solve a problem or answer a significant question based on the lesson; or to give the students the opportunity to ask questions they still have about the lesson; or to see if students can apply the lesson they learned in a new way. In the \textit{collection} step, teachers allocate a specific amount of time for students to complete the survey at the end of the class before they leave. In the \textit{analysis} step, teachers go through the surveys one by one and analyze the responses by extracting useful pieces of information out of them. Finally, in the \textit{action} step by relying on the analysis of the responses, the teachers make certain adjustments in their upcoming lectures in order to take into account the feedback they directly gathered from the students. In the action step, for example, teachers may decide to review the concepts that students had difficulties understanding, or they may consider changing their teaching style or the teaching tools they use. \\
\indent Teachers commonly use four types of questions in surveys, namely, (i) multiple-choice, (ii) rating scale, (iii) likert scale, and (iv) open-response. In the analysis step, processing the responses of multiple-choice, rating scale, and likert scale questions is straightforward as the expected set of responses are known ahead of time and are limited to the choices that teachers make during the creation step. However, analyzing the responses to open-response questions is not straightforward. This is due to the fact that the open-responses can include a wide range of topics raised by the students and such topics can significantly vary from one student to another. For analyzing the open-responses teachers commonly go through the responses one by one and try to read all of them once. After getting a rough idea about the theme of the topics conveyed through the responses of students, teachers narrow down the topics and see which student is talking about what topic in order to get a general overview of the discussed topics across the class. This approach for analyzing the open-responses is extremely time-consuming, inefficient, challenging, and often biased. \\
\indent Surveys in their principles are identical to surveys that are commonly used in different applications in order to directly gather feedback from customers or users. There have been efforts to automate survey analysis in order to save time as well as cost and be able to use the outcome of such analysis to increase user satisfaction and improve services. A common approach for analyzing open-ended survey responses is the use of topic modeling \cite{Vayansky2020}. In machine learning and natural language processing domains, topic modeling is a popular text-mining technique that upon scanning a set of documents detects textual patterns within them and automatically clusters word groups and similar expressions that best characterize a collection of documents. The three most common techniques of topic modeling are, namely, (i) Latent Semantic Analysis (LSA), (ii) probabilistic Latent Semantic Analysis (pLSA), and (iii) Latent Dirichlet Allocation (LDA) \cite{Alghamdi2015}. The main challenge behind such commonly used topic modeling techniques is that they do not account for contextual information and they merely rely on the word-level frequencies and even ignore the order and semantic relationships between the words and sentences. Moreover, such context-agnostic topic modeling approaches require careful pre-processing of textual data, are vulnerable to overfitting due to not generalizing well on unseen text samples, need extensive hyperparameter tuning, and cannot capture implicit semantics such as sarcasm, anger, humor, metaphors, and figurative languages \cite{Buenano-Fernandez2020,Pietsch2018,Finch2018}. In order to overcome such shortcomings, in this work, we present an end-to-end framework for context-aware analysis of survey open-responses which can also be directly used for analyzing any other type of open-response survey data. \\
\indent The remainder of this paper is organized as follows. In section \eqref{sec:methodology}, we present the details of the methodology behind context-aware analysis of open-responses. We discuss context-aware clustering in subsection \eqref{sec:un-supervised-clustering}, context-aware cluster assignment in subsection \eqref{sec:semi-supervised-clustering}, and context-aware insights in subsection \eqref{sec:context-aware-insights}. In section \eqref{sec:dataset}, we present an overview of the dataset used in this work. In section \eqref{sec:results}, we present the results and discussions. Finally, in section \eqref{sec:conclusion}, we present the concluding remarks as well as some insights on future directions.
\begin{figure}[!h]
    \centerline{\includegraphics[width=0.78\paperwidth]{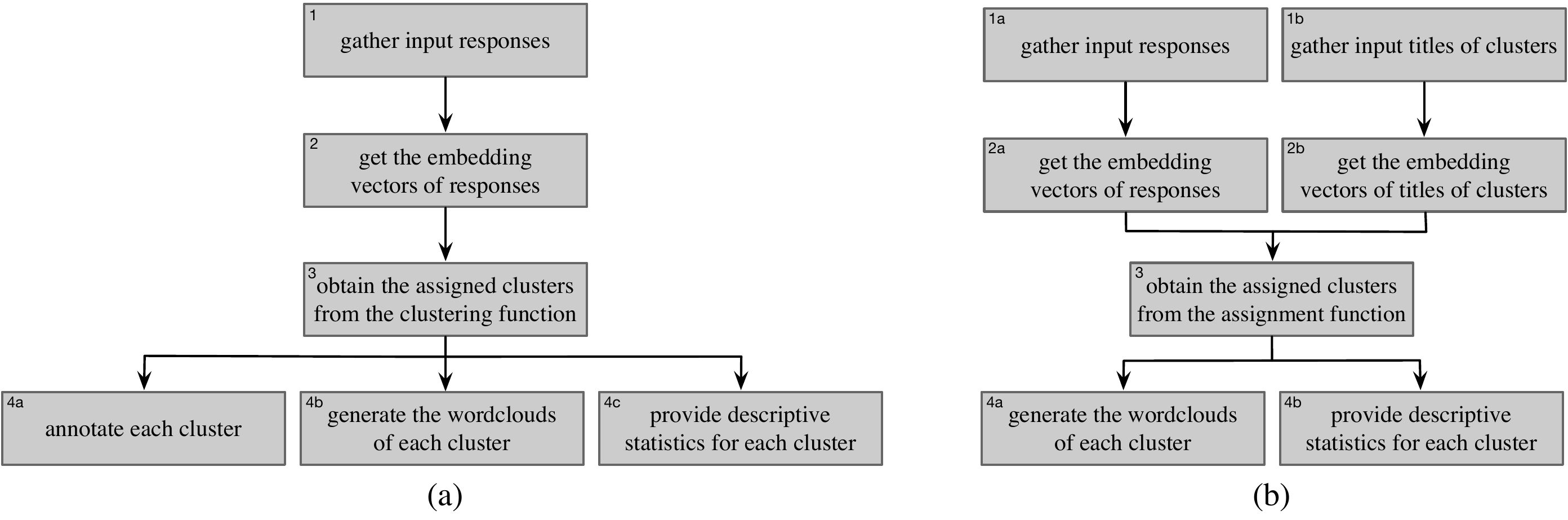}}
    \caption{Overview of the steps in (a) context-aware clustering and (b) context-aware cluster assignment. In context-aware clustering (a), we first gather a list of open-responses submitted by students. Next, using the pre-trained language model we generate the embedding vectors of the responses. Afterwards, we obtain the clusters of responses from the clustering function. Finally, we annotate each cluster with an appropriate title (\ie prominent keywords), generate wordclouds, and provide descriptive statistics for each cluster. In context-aware cluster assignment (b), in the first step, in addition to gathering the list of open-responses we also gather a list of labels of clusters as the input. Next, using the pre-trained language model we generate the embedding vectors for the responses as well as the labels of the clusters. Afterwards, we obtain the clusters of responses from the assignment function. Finally, since the cluster labels are already provided as input, we only generate wordclouds, and provide descriptive statistics for each cluster at the end.}
    \label{fig:framework_flowchart}
\end{figure}

\section{Methodology} \label{sec:methodology}
\noindent In this work, we use pre-trained natural language models in order to extract the contextual semantic patterns in a collection of open-responses. Pre-trained natural language models \cite{Qiu2020} are commonly used for a wide range of tasks such as text generation \cite{Li2021}, building dialogue systems \cite{Budzianowski2019}, text classification \cite{Kant2018}, hate speech detection \cite{Melo2019}, sentiment analysis \cite{Araci2019}, named entity recognition \cite{Giorgi2019}, question answering \cite{Pearce2021}, and text summarization \cite{Kryscinski2020,1904.00788}. \\
\indent There are two main categories of pre-trained language models, namely, (i) word-level models and (ii) sentence-level models. Pre-trained word-level models such as Word2Vec \cite{Mikolov2013} and GloVe \cite{pennington2014glove} are used to encode words into so-called \textit{embedding} vectors. One major limitation of word-level embedding vectors obtained by Word2Vec and GloVe is that they do not capture the context of the words in sentences. Context-aware word-level embeddings such as ELMo \cite{Peters2018} and BERT \cite{Devlin2019} attempt to address that shortcoming by accounting for the context of a word within a sentence. Recently, pre-trained language models have been used for capturing the contexts beyond word-level in order to encode sentences into embedding vectors. Commonly used pretained language models for getting sentence-level embeddings are Sentence-BERT (SBERT) \cite{Reimers2020}, GPT \cite{Radford}, GPT-2 \cite{Radford2020}, RoBERTa \cite{Liu2019}, XLNet \cite{Yang2019}, DistilBERT \cite{Sanh2019}, Transformer-XL \cite{Dai2020a}, InferSent \cite{Conneau2017}, Universal Sentence Encoder \cite{Wang2010}, and Skip-Thoughts \cite{Kiros}. In this work, we use the Sentence-BERT (SBERT) \cite{Reimers2020} model provided by HuggingFace \cite{WinNT} to get the embedding vectors of words as well as sentences. 
\subsection{Context-Aware Clustering} \label{sec:un-supervised-clustering}
\noindent In this section, we present an overview of our clustering approach. In the clustering task, the inputs are the raw open-responses gathered from the surveys. Using the SBERT pre-trained language model we first tokenize and then extract the embedding vectors for each input sample. Basically, the SBERT model maps sentences and paragraphs to a $384$ dimensional dense vector space that can be used for downstream tasks such as clustering or semantic search. Once the embedding vectors are created we use the k-means algorithm \cite{Hartigan1979} to cluster the input samples. In the k-means algorithm, the number of clusters $k$ is unknown and needs to be found and provided as an input. We use the silhouette score (see, Appendix \eqref{sec:silhouette-score}) to find the best number of clusters $k^*$. In order to find $k^*$ we explore the silhouette score values for multiple number of clusters between $2$ and an upper bound of $k_{max}$ and choose $k^*$ as the number of clusters where the silhouette score obtains its maximum value as
\begin{equation}\label{eq:solhoueteScore}
	k^* = \text{arg}\max_{k} SS^{(k)}\,\,\, \text{for}\,\,k\in [2,\,k_{max}]\,,
\end{equation}
where $SS^{(k)}$ is the silhouette score for a given clustering configuration with $k$ number of clusters. Finally, we annotate each cluster with the prominent keywords of its samples (section \eqref{sec:clusterAnnotation}), we generate wordclouds for each cluster as well as a unified wordcloud for all the samples (section \eqref{sec:wordcloudGeneration}), and we provide descriptive statistics for each cluster (section \eqref{sec:providingStatistics}). Figure (\ref{fig:framework_flowchart}a) shows an overview of the steps involved in the context-aware clustering process.
\subsection{Context-Aware Cluster Assignment} \label{sec:semi-supervised-clustering}
\begin{figure}[!h]
    \centerline{\includegraphics[width=0.82\paperwidth]{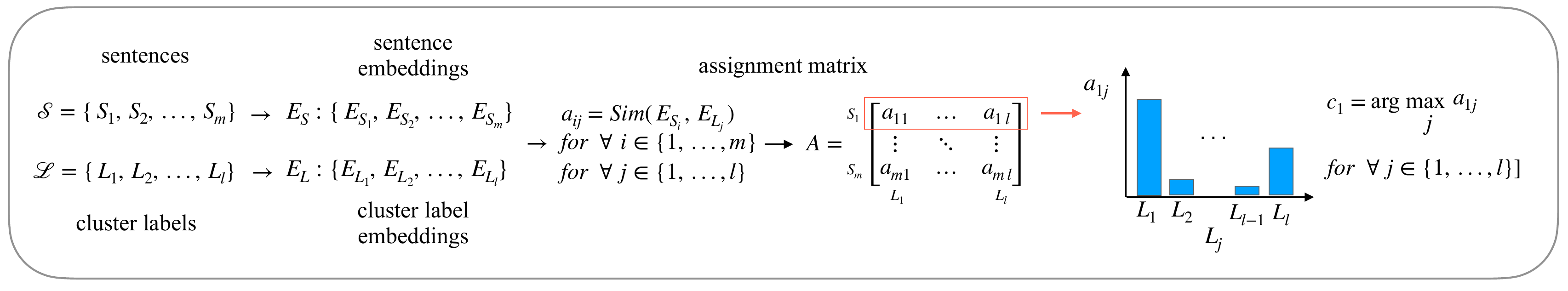}}
    \caption{Overview of the cluster assignment approach when sentences ($\mathcal{S}$) and clusters labels ($\mathcal{L}$) are provided as inputs. For all the sentences and cluster labels the corresponding set of embeddings (\ie $E_S$ and $E_L$) get generated by the pre-trained language model. The matrix $A$ designates the so-called assignment matrix where each element of it ($a_{ij}$) corresponds to the cosine similarity ($Sim$) between the embedding pairs of sentences ($E_{S_i}$) and clusters labels ($E_{Lj}$). Finally, as an example, the assigned cluster to sentence $S_1$ is the most cosine similar label in the embedding space.}
    \label{fig:cluster_assignment}
\end{figure}
\noindent In this section, we present an overview of our clustering assignment approach. In the clustering assignment approach, we have two groups of inputs, namely, the raw open-responses ($\mathcal{S}$) and the labels of the clusters ($\mathcal{L}$). Using the SBERT pre-trained language model we tokenize and extract the embedding vectors $E_S$ and $E_L$, respectively for each group of inputs. Next, we calculate the assignment matrix $A$ with its elements being the pairwise cosine similarity between the sentence embeddings and the cluster label embeddings as
\begin{equation}
A = a_{ij} = Sim(E_{S_i}, E_{L_j})\,\,\,\,\text{for}\,\, \forall i \in \lbrace{1,\ldots,m\rbrace},\,\,\text{for}\,\, \forall j \in \lbrace{1,\ldots,l\rbrace}\,,
\end{equation}    
where $m$ and $l$ are the number of input open-responses and input cluster labels, respectively. $Sim(\,)$ represents the cosine similarity function and for two vectors $\mathbf{u}$ and $\mathbf{v}$ it can be defined as
\begin{equation}
Sim(\mathbf{u}, \mathbf{v}) = \frac{\mathbf{u}\cdot\mathbf{v}}{\|\mathbf{u}\|\|\mathbf{v}\|} = \frac{\sum\limits_{i=1}^{V} u_iv_i}{\sqrt{\sum\limits_{i=1}^{V} u^2_i}\sqrt{\sum\limits_{i=1}^{V} v^2_i}}\,,
\end{equation}
where $V$ represents the length of the vectors. In this work, the length of the embedding vectors generated by the SBERT models is $V=384$. Once we build the assignment matrix $A$, the corresponding assigned label $c_i$ to a sentence $i$ can be found as
\begin{equation}
c_i = \text{arg}\max_ja_{ij}\,\,\,\,\text{for}\,\,\forall j \in \lbrace{1,\ldots,l\rbrace},\,\,\text{for}\,\, \forall i \in \lbrace{1,\ldots,m\rbrace}\,.
\label{eq-cluster-assignment}
\end{equation}
In equation \eqref{eq-cluster-assignment} each sentence is assigned the label with the highest value of cosine similarity in the embedding space. Figure \eqref{fig:cluster_assignment} illustrates the steps involved in the cluster assignment approach for calculating the assignment matrix $A$ and finding the assigned labels. Finally, we provide a set of descriptive statistics for each cluster (section \eqref{sec:providingStatistics}). It is worth noting that since the cluster labels are provided as an input in the cluster assignment approach, different from clustering in section \eqref{sec:un-supervised-clustering}, we do not annotate the clusters. The input titles are basically considered as the annotation of the clusters. Figure (\ref{fig:framework_flowchart}b) shows an overview of the steps involved in the context-aware cluster assignment process.

\subsection{Context-Aware Insights}\label{sec:context-aware-insights}

\begin{figure}[!h]
    \centerline{\includegraphics[width=0.82\paperwidth]{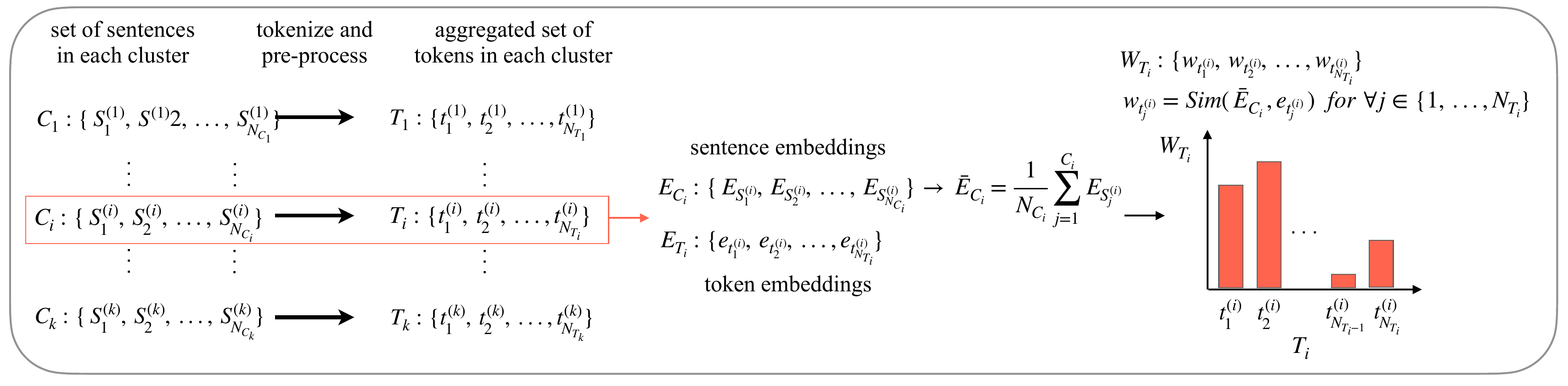}}
    \caption{Overview of the cluster annotation approach, \ie labeling the clusters. Upon clustering we obtain the sentences within each cluster $C_i$. For every cluster $C_i$, we extract and pre-process the set of all tokens $T_i$ out of the sentences. Next, for each sentence $S^{(i)}_j$ and every token $t^{(i)}_j$ of cluster $C_i$ we generate their corresponding embedding representations, \ie $E_{S^{(i)}_j}$ and $e_{t^{(i)}_j}$, respectively. Next, we calculate the average value ($\bar{E}_{C_i}$) of the set of sentence embeddings ($E_{C_i}$), where $N_{C_i}$ is the number of sentences in cluster $C_i$. Afterward, for cluster $C_i$ we calculate the weight value ( $w_{t^{(i)}_j}$) of each token $t_j$ as the cosine similarity ($Sim$) between $\bar{E}_{C_i}$ and its embedding vector $e_{t^{(i)}_j}$ where $N_{T_i}$ represents the number of tokens in cluster $C_i$.}
    \label{fig:tokenization_weights}
\end{figure}

\subsubsection{Cluster Annotation:} \label{sec:clusterAnnotation}

\noindent In this part, we explain our approach for context-aware annotation of the clusters in the clustering task presented in section \eqref{sec:un-supervised-clustering}. In the clustering task, the output of the clustering function is a set of grouped input open-responses where each group ($C_i$) designates an independent cluster. Assuming that the cluster $C_i$ has $N_{C_i}$ number of sentences, we tokenize all the sentences and carry out a few pre-processing steps on the tokens. The pre-processing steps mainly include removing the stop-words, lemmatization and/or stemming. After the tokenization and post-processing steps, we gather the set of tokens $T_i$ corresponding to cluster $C_i$. Next, using the SBERT pre-trained language model we extract the embedding vectors for the set of sentences ($E_{C_i}$) and tokens ($E_{T_i}$), respectively. Afterward, we calculate the average value of the set of sentence embeddings as $\bar{E}_{C_i}$, which is the centroid of a cluster in the embedding space. Next, for each cluster $C_i$, we calculate the weight value $w_t$ of each token as the cosine similarity between $\bar{E}_{C_i}$ and the token embedding vector $e_t$. We then sort the set of tokens in each cluster $C_i$ with respect to the $w_t$ values in descending order. Then we pick the top $5$ prominent tokens (\ie 5 tokens with the largest $w_t$ values) and use them for annotating each cluster. Figure \eqref{fig:tokenization_weights} illustrates an overview of the described cluster annotation approach.
\subsubsection{Wordcloud Generation:} \label{sec:wordcloudGeneration}
\noindent Wordcloud (aka. tag cloud) \cite{5290722} is a visual representation of text data that is commonly used to depict keywords where the sizes of the words represent their frequency or importance level. In this work, we present an approach for creating context-aware wordclouds. We consider two different types of wordclouds, namely, (i) cluster-level wordclouds for each cluster, and (ii) unified wordcloud for all the input open-responses. The cluster-level wordclouds include the top prominent tokens with their sizes corresponding to the weight values found in subsection \eqref{sec:clusterAnnotation}. The unified wordcloud on the other hand characterizes the prominent keywords across all the input open-responses. In order to create the unified wordcloud, we use the words in the cluster-level wordclouds together with their corresponding weight values ($w$). Next, we scale the weight values with a density coefficient ($\rho$) that accounts for the relative number of samples in each cluster. Density coefficient $\rho_i$ for cluster $C_i$ is defined as
\begin{equation}
	\rho_i = N_{C_i}/m\,\,\,\text{where}\,\,i\in\lbrace{1,\ldots,k\rbrace}\,,
\end{equation}
where $N_{C_i}$ is the number of samples in cluster $C_i$, $m$ is the total number of input open-responses, and $k$ is the number of clusters. Accordingly, the scaled weight values of the prominent tokens ($w^*$) for the cluster $C_i$ become
\begin{equation}
	w^*_{t^{(i)}_j} = \rho_iw_{t^{(i)}_j} \,\,\,\,\text{for}\,\,\forall j \in \lbrace{1,\ldots,5\rbrace},\,\,\text{for}\,\, \forall i \in \lbrace{1,\ldots,k\rbrace}\,,
\end{equation}
where $k$ is the number of clusters. It is worth noting that the scaled weight values ($w^*$) not only take into account the importance of each keyword at the cluster level (\ie through $w$), but also account for the relative importance of each cluster depending on how many samples a cluster entails (\ie through $\rho$).
\subsubsection{Descriptive Statistics:} \label{sec:providingStatistics}
\noindent After doing clustering or cluster assignment, we can provide descriptive statistics about the survey responses. One insight that we can provide upon doing clustering is the number of clusters (groups) that the input responses fall under. Upon doing clustering or cluster assignment another descriptive statistic that we can provide is the number (ratio) of samples that fall under each category. We can further provide a similarity matrix that captures the similarity between all the cluster pairs by calculating the cosine similarity between their centroids in the embedding space. Moreover, using the samples in each cluster we can provide certain insights. For instance, the minimum, maximum, and average number of words in each cluster can be provided as a cluster-level insight. Another cluster-level insight is the sentimental distribution of the samples in each cluster, where sentiments could for instance be positive/negative/neutral or happy/sad or any other sentiment groups that we can provide using a sentiment analysis subroutine. \\
\indent As the above descriptive insights are provided for open-responses in surveys, one valuable piece of information that can be presented is that how they evolve over time. For example in case the same set of survey questions get repeated during a semester, how do the cluster distributions, their annotations, their wordclouds and etc. evolve. By looking at such evolutions, we can provide feedback on how the set of actions that were taken by the survey owners (\eg teachers) based on the previous surveys have affected the opinions of responders over time.

\section{Dataset}\label{sec:dataset}
\noindent As the dataset for this work, the authors of this manuscript have manually written a plausible set of responses to an open-response survey question for a chemistry class in response to an open-response survey question. In this survey the teacher asks about one topic that each student would like to be reviewed before their upcoming exam. The important point to note here is that asking these type of open-ended questions in surveys cannot be done through multiple-choice type questions for instance. Hence, an open-response question is the most appropriate format. As mentioned earlier, processing the responses of such open-ended questions is costly and hard. Accordingly, in this work, we will show how we can efficiently process the responses to such open-ended survey questions. Table \eqref{table:inputresponses} shows the list of responses.

\begin{table}[!h]
\centering
\arrayrulecolor{black}
\resizebox{0.85\columnwidth}{!}{
\begin{tabular}{|l|l|} 
\hline
\rowcolor{black} \textcolor{white}{ID} & \multicolumn{1}{c|}{\textcolor{white}{Input Responses of Students}}                                                \\ 
\hline
\#1  & About acids \& bases that we learned in the last lecture. \\ 
\#2  & About how we can use periodic table to identify reactions.            \\ 
\#3  & About the difference between bases and acids in their chemical formula.            \\ 
\#4  & About the differences between entropy and enthalpy, and also their similarities.       \\ 
\#5  & About the differences between ionic bonding \& covalent bonding.             \\ 
\#6  & About the differences between proton and neutron.         \\ 
\#7  & About the ionic bonding and its properties when reacting with other substances.         \\ 
\#8  & About the similarities between protons and neutrons with electrons.           \\ 
\#9  & About the total number of elements in the periodic table that we studied.            \\ 
\#10 & About the use cases of both acids and bases in industry.          \\ 
\#11 & About the way atoms join together through ionic and covalent bonding.            \\ 
\#12 & Could you please explain how entropy can get transformed in to enthalpy and vice versa.         \\ 
\#13 & Explain the entropy and enthalpy concepts that we learned in the beginning of semester.           \\ 
\#14 & I have a hard time understanding how in ionic bonding, atoms transfer electrons to each other?            \\ 
\#15 & Please elaborate on the reactions of acids and bases with inert compounds.          \\ 
\#16 & Please explain the applications of acids in chemistry.          \\ 
\#17 & Regarding the periodic table and the order of chemical elements in each column.          \\ 
\#18 & Why entropy and enthalpy are important and how they are used in thermodynamic.      \\ 
\#19 & about how distance unit in foot can get converted to distance unit in meter.          \\ 
\#20 & about the unit conversion in SI system and how that differs with UK system.        \\ 
\#21 & could you please explain why Ionic bonds form between a metal and a nonmetal mostly?            \\ 
\#22 & it would be great if you could explain more about the atomic structure and neutron, proton, electron.            \\ 
\#23 & please clarify about the covalent and ionic bonding and how they are different and similar.         \\ 
\#24 & please explain about units and how unit conversion works.    \\ 
\#25 & please explain how we can transform pounds unit to kilograms.       \\ 
\#26 & please explain more about the composition of atoms such as electron, neutron and proton.      \\ 
\#27 & please explain unit conversion again with a few more examples.     \\ 
\#28 & please explain what are the common properties of protons and neutrons. \\
\arrayrulecolor{black}\hline
\end{tabular}
}
\caption{List of a plausible set of responses to an open-response survey question for a chemistry class that the authors of this manuscript have manually written for an open-response survey question where the teacher asks about one topic that the students would like to be reviewed again before their upcoming exam.}
\label{table:inputresponses}
\end{table}

\section{Results and Discussion} \label{sec:results}
\noindent In this section, we present the results for context-aware clustering (subsection \eqref{sec:unsup_res}) as well as context-aware cluster assignment (subsection \eqref{sec:semisup_res}) tasks. In the latter case, we consider a set of cluster topics provided by the teacher as input.
\subsection{Context-Aware Clustering}\label{sec:unsup_res}
\begin{figure}[!h]
    \centerline{\includegraphics[width=0.65\paperwidth]{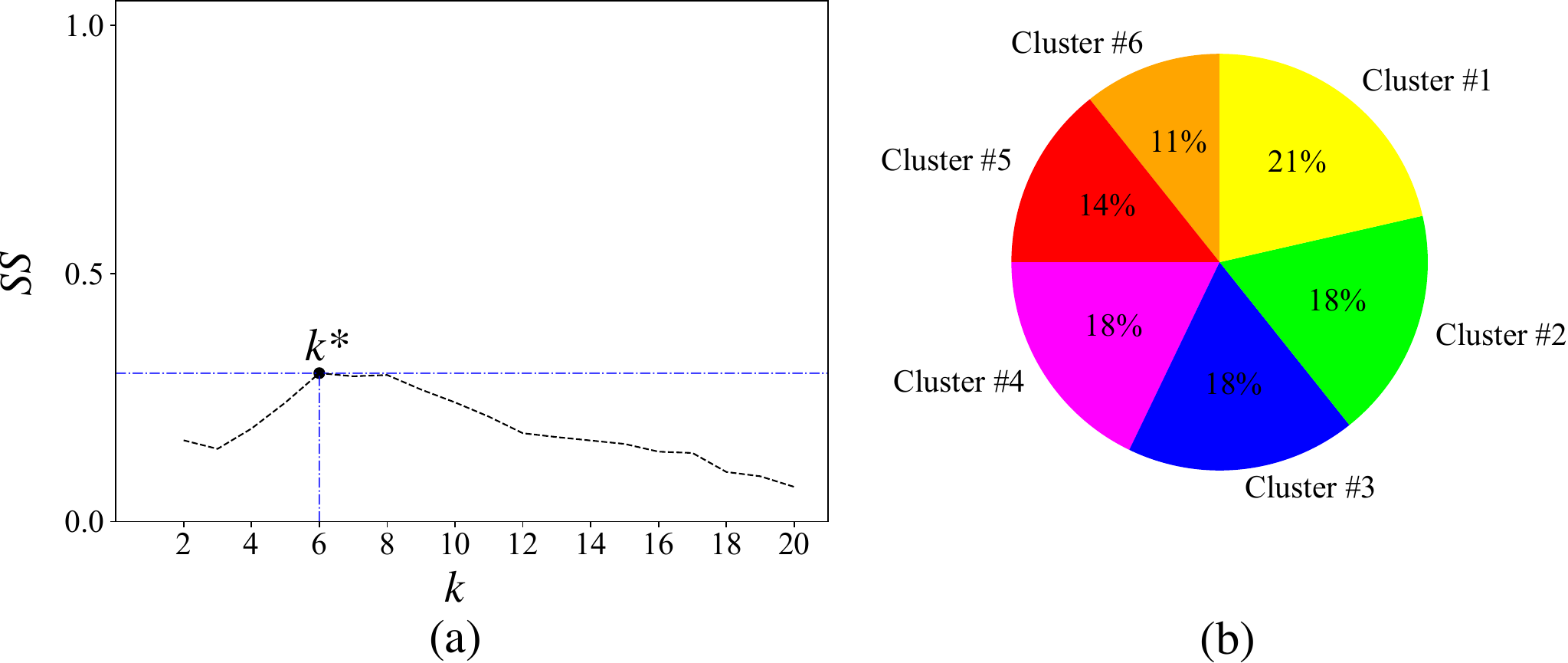}}
    \caption{(a) The silhouette score ($SS$) as a function of the number of clusters ($k$) for $k\in \lbrace{ 2, \ldots, 20\rbrace}$. The best number of clusters is $k^*=6$ where the silhouette score obtains its maximum value (see, equation \eqref{eq:solhoueteScore}). (b) The pie chart that shows the distribution of the number of samples ($N_C$) that belong to each cluster ($i$) where $N_{C_i} = \lbrace{ 6, 5, 5, 5, 4, 3 \rbrace}$ for $i \in \lbrace{ 1, 2, 3, 4, 5, 6 \rbrace}$.}
    \label{fig:silhouette_vals_vs_nclusters}
\end{figure}

\begin{figure}[!h]
    \centerline{\includegraphics[width=0.75\paperwidth]{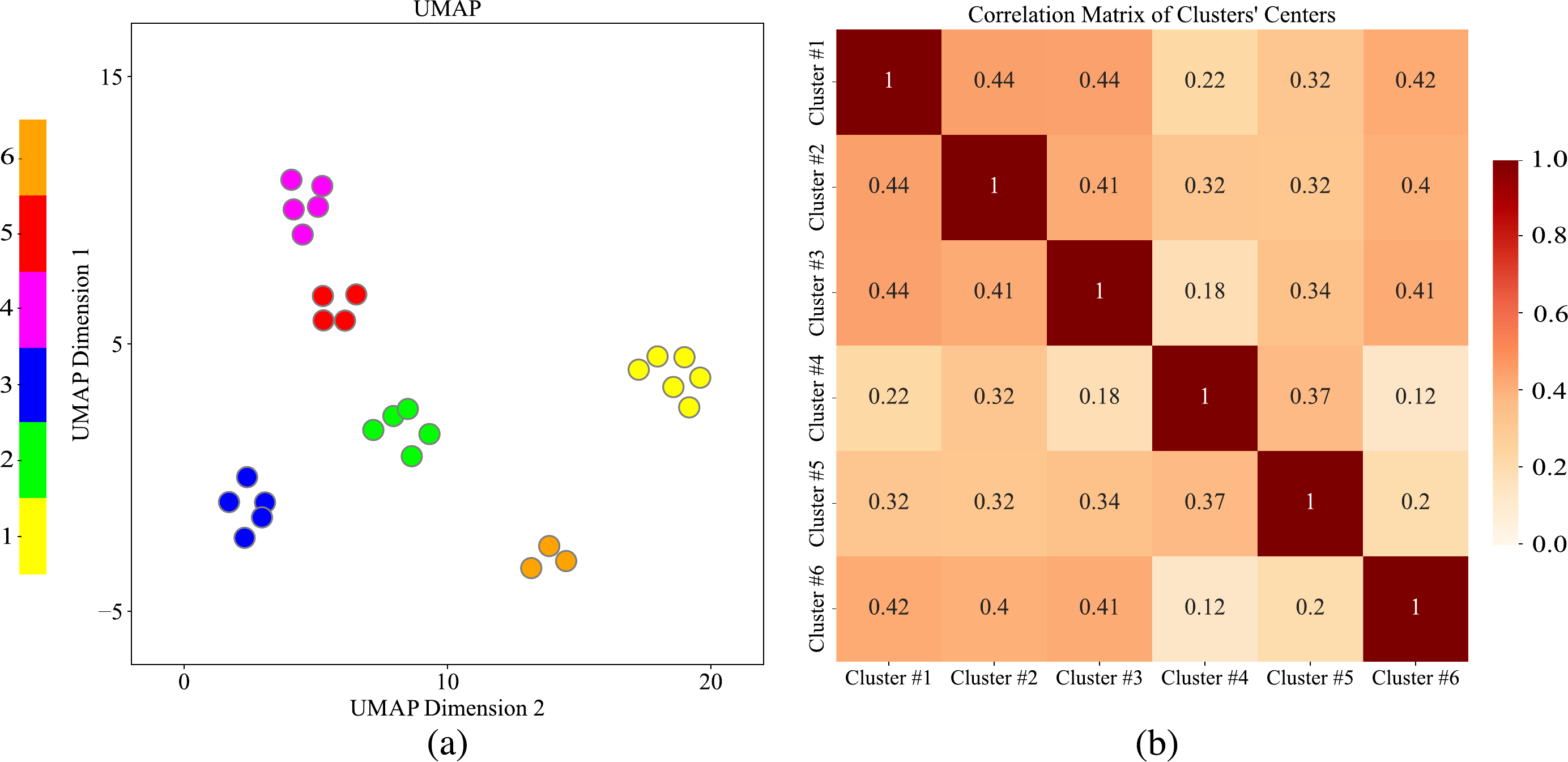}}
    \caption{(a) Application of UMAP \cite{McInnes2018} to the sentence embeddings where the colors represent the clusters. (b) Correlation matrix of clusters' centers where each entry represents the cosine similarity value between the centroids in a pair of two clusters.}
    \label{fig:ConfMatrix_Umap}
\end{figure}

\begin{figure}[!h]
    \centerline{\includegraphics[width=0.8\paperwidth]{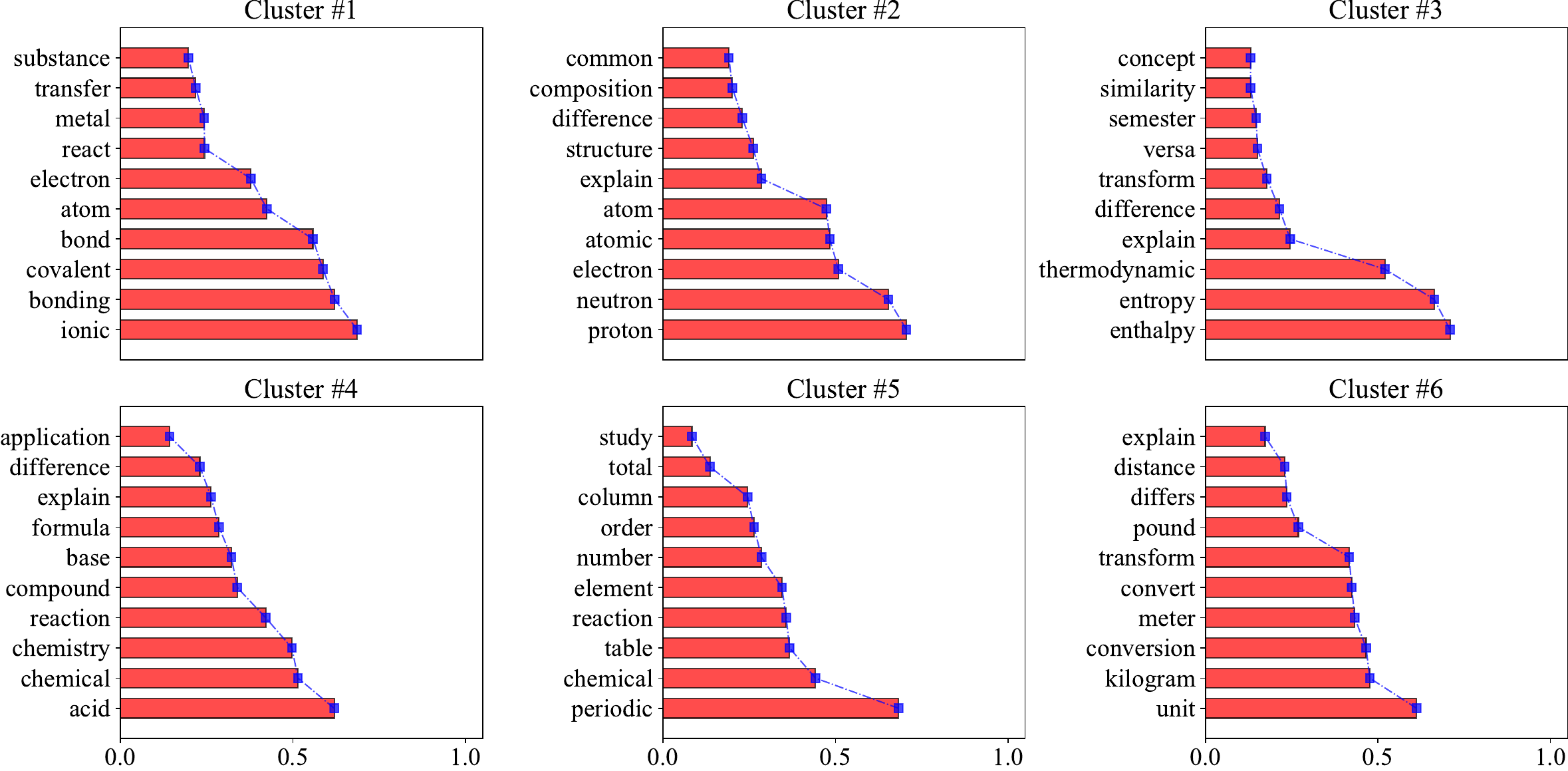}}
    \caption{The tokens importance values ($W_{T_i}$) for the top 10 tokens of each cluster $C_i$ is illustrated where $i \in \lbrace{ 1, 2, 3, 4, 5, 6 \rbrace}$.}
    \label{fig:tokens_importance}
\end{figure}

\noindent Figure (\ref{fig:silhouette_vals_vs_nclusters}a) shows the value of silhouette score ($SS$) as a function of the number of clusters ($n$) for $n\in \lbrace{ 2, \ldots, 20\rbrace}$. We can see that for few (\ie $k\rightarrow 2$) as well as large (\ie $k\rightarrow 20$) number of clusters the silhouette score values drop. As we can see, the maximum value of the silhouette score happens for $k^*=6$ number of clusters where $SS^{(k^*)}=0.299$. Accordingly, in context-aware clustering, we consider the number of clusters to be six. As a descriptive statistic, figure (\ref{fig:silhouette_vals_vs_nclusters}b) shows a pie chart that illustrates the distribution of the number of samples ($N_C$) that belong to each cluster ($i$) with the number of samples in each cluster being $N_{C_i} = \lbrace{ 6, 5, 5, 5, 4, 3 \rbrace}$ for $i \in \lbrace{ 1, 2, 3, 4, 5, 6 \rbrace}$. We see that the majority of samples fall under cluster $C_1$ and cluster $C_6$ has the smallest number of samples associated with it. 
\\
\indent Table \eqref{tab:clusters} shows the list of clustered responses distributed across the six clusters $C_1$ to $C_6$. If we look at the responses categorized under each cluster, we can see that the responses that have fallen under each group discuss identical topics. For instance, cluster $C_1$ entails responses that mostly talk about atomic interactions and molecular bondings, whereas cluster $C_6$ for instance includes the questions asked by students about different forms of unit conversion.
\begin{table}[!h]
\centering
\arrayrulecolor{black}
\resizebox{0.9\columnwidth}{!}{
\begin{tabular}{|c|l|} 
\hline
\rowcolor{black} \textcolor{white}{Cluster ID} & \multicolumn{1}{c|}{\textcolor{white}{Clustered Responses}}                                             \\ 
\hline
\multirow{6}{*}{$C_1$}                 & About the differences between ionic bonding \& covalent bonding.  \\ 
\arrayrulecolor[rgb]{0.753,0.753,0.753}
\cline{2-2}
                                             & About the ionic bonding and its properties when reacting with other substances. \\ 
\cline{2-2}
                                             & About the way atoms join together through ionic and covalent bonding. \\ 
\cline{2-2}
                                             & I have a hard time understanding how in ionic bonding, atoms transfer electrons to each other? \\ 
\cline{2-2}
                                             & could you please explain why Ionic bonds form between a metal and a nonmetal mostly? \\ 
                                             
\cline{2-2}
                                             & please clarify about the covalent and ionic bonding and how they are different and similar.\\                 
\arrayrulecolor{black}                                             
\hline
\multirow{5}{*}{$C_2$}                    & About the differences between proton and neutron. \\ 
\arrayrulecolor[rgb]{0.753,0.753,0.753}
\cline{2-2}
                                               & About the similarities between protons and neutrons with electrons. \\ 
\cline{2-2}
                                               & it would be great if you could explain more about the atomic structure and neutron, proton, electron. \\ 
\cline{2-2}
                                               & please explain more about the composition of atoms such as electron, neutron and proton. \\ 
\cline{2-2}
\arrayrulecolor{black}                                               & please explain what are the common properties of protons and neutrons. \\ 
\hline
\multirow{4}{*}{$C_3$}                   & About the differences between entropy and enthalpy, and also their similarities. \\ 
\arrayrulecolor[rgb]{0.753,0.753,0.753}
\cline{2-2}
                                               & Could you please explain how entropy can get transformed in to enthalpy and vice versa. \\ 
\cline{2-2}
                                               & Explain the entropy and enthalpy concepts that we learned in the beginning of semester. \\ 
\cline{2-2}
                                            & Why entropy and enthalpy are important and how they are used in thermodynamic. \\
\arrayrulecolor{black}                                              
\hline
\multirow{5}{*}{$C_4$}                    & About acids \& bases that we learned in the last lecture. \\ 
\arrayrulecolor[rgb]{0.753,0.753,0.753}
\cline{2-2}
                                               & About the difference between bases and acids in their chemical formula.  \\ 
\cline{2-2}
                                               & About the use cases of both acids and bases in industry.\\ 
\cline{2-2}
                                               & Please elaborate on the reactions of acids and bases with inert compounds. \\ 
\cline{2-2}
\arrayrulecolor{black}                                               & Please explain the applications of acids in chemistry.\\ 
\hline
\multirow{3}{*}{$C_5$}                    & About how we can use periodic table to identify reactions.\\ 
\arrayrulecolor[rgb]{0.753,0.753,0.753}
\cline{2-2}
                                               & About the total number of elements in the periodic table that we studied. \\ 
\cline{2-2}
                                             & Regarding the periodic table and the order of chemical elements in each column.\\ 
\arrayrulecolor{black}  
\hline
\multirow{5}{*}{$C_6$}                    & about how distance unit in foot can get converted to distance unit in meter. \\ 
\arrayrulecolor[rgb]{0.753,0.753,0.753}
\cline{2-2}
                                               & about the unit conversion in SI system and how that differs with UK system. \\ 
\cline{2-2}
                                               & please explain about units and how unit conversion works. \\ 
\cline{2-2}
                                               & please explain how we can transform pounds unit to kilograms. \\ 
\cline{2-2}
                                               & please explain unit conversion again with a few more examples.\\ 
\arrayrulecolor{black}
\hline
\end{tabular}
}
\arrayrulecolor{black}
\caption{List of clustered responses distributed across the six clusters $C_1$ to $C_6$.}
\label{tab:clusters}
\end{table}
\\
\indent Figure (\ref{fig:ConfMatrix_Umap}a) illustrates the UMAP \cite{McInnes2018} projection of the sentence embeddings where the color represents different clusters. We can see that through the UMAP projection in the embeddings space the clusters get properly segregated and form six distinct groups. Figure (\ref{fig:ConfMatrix_Umap}b) shows the correlation matrix for the centers of clusters where each entry represents the cosine similarity value between the centroids in a pair of two clusters. We can see that the cosine similarity between the centroids of any pair of two clusters is always below 0.5. Building such a correlation matrix helps to understand how distinct the clusters are and highlight when two clusters are similar (\ie their cosine similarity is above a threshold value \eg 0.8). In this case, we can either merge them or reduce the number of clusters and do re-clustering. As this correlation-based approach considers how segregated the centroids of the clusters are with respect to one another, it can be used as an additional way of inspecting the clustering quality besides merely choosing the number of clusters as the one corresponding to the maximum Silhouette score.
\\
\indent Next, we annotate each cluster by following the steps presented in subsection \eqref{sec:clusterAnnotation}. Figure \eqref{fig:tokens_importance} shows the importance values ($W_{T_i}$) of top 10 tokens for each cluster $C_i$ where $i \in \lbrace{ 1, 2, 3, 4, 5, 6 \rbrace}$. We can see that the top five tokens for clusters $C_1$ to $C_6$ respectively are: \{ionic, bonding, covalent, bond, atom\}, \{proton, neutron, electron, atomic, atom\}, \{enthalpy, entropy, thermodynamic, explain, difference\}, \{acid, chemical, chemistry, reaction, compound\}, \{periodic, chemical, table, reaction, element\}, \{unit, kilogram, conversion, meter, convert\}.

\begin{figure}[!h]
    \centerline{\includegraphics[width=0.7\paperwidth]{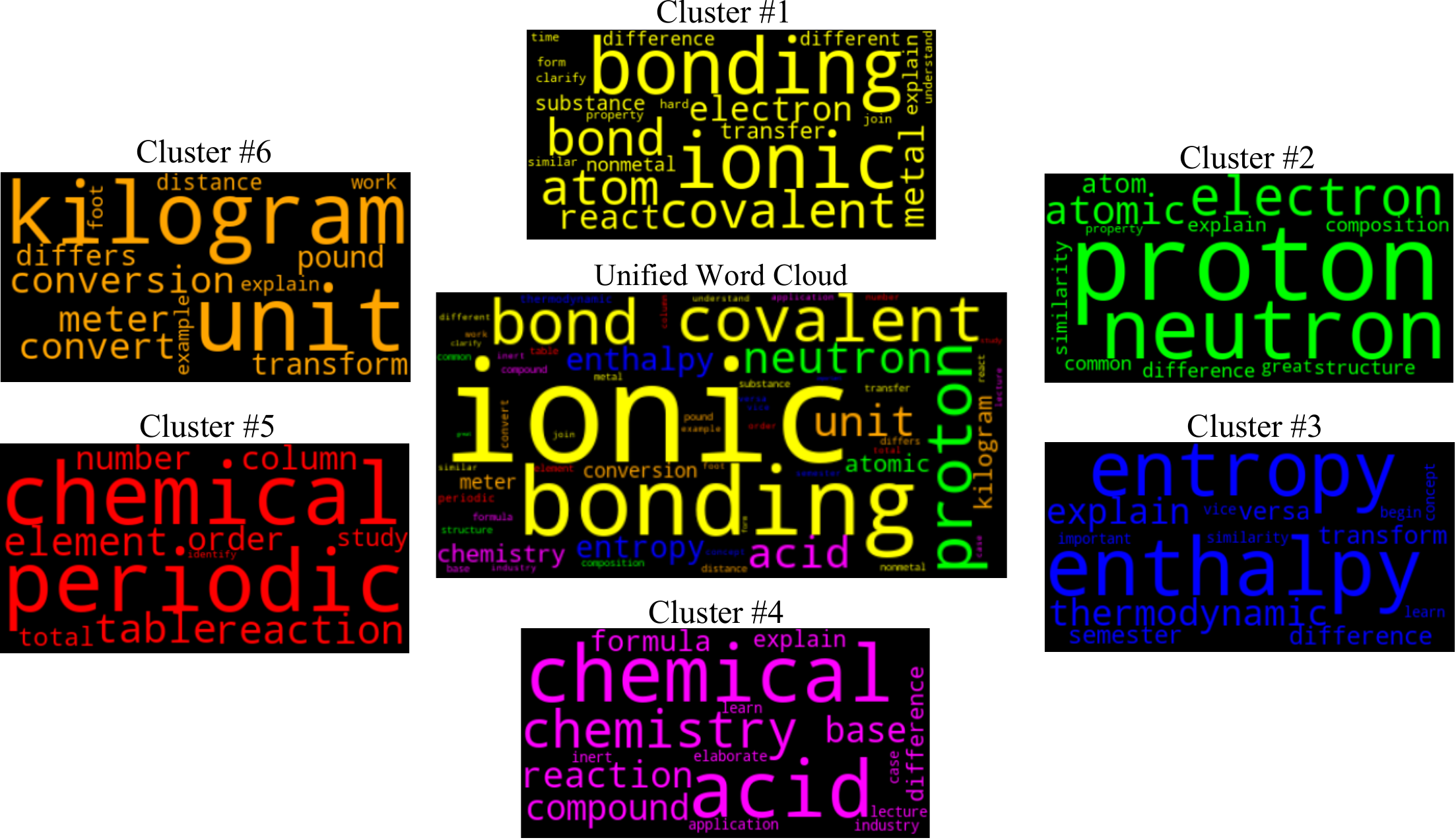}}
    \caption{Wordclouds for the prominent tokens of each cluster $C_i$ for $i \in \lbrace{ 1, 2, 3, 4, 5, 6 \rbrace}$ are illustrated where the size of words represents the tokens importance values ($W_{T_i}$) in each cluster. The middle wordcloud shows the unified wordcloud generated using the approach presented in section \eqref{sec:wordcloudGeneration} where the colors represent the association of words to each cluster and the size of the words represent the scaled weight values of the prominent tokens ($w^*$).}
    \label{fig:all_wordClouds}
\end{figure}

\indent Figure \eqref{fig:all_wordClouds} shows the wordclouds for the prominent tokens of each cluster $C_i$ for $i \in \lbrace{ 1, 2, 3, 4, 5, 6 \rbrace}$ where the size of words represents the tokens importance values ($W_{T_i}$) in each cluster. Wordcloud of each cluster is designated with a different color consistent with the colors used in Figures (\ref{fig:silhouette_vals_vs_nclusters}b) and (\ref{fig:ConfMatrix_Umap}a). In figure \eqref{fig:all_wordClouds} the middle wordcloud shows the unified wordcloud generated using the approach presented in section \eqref{sec:wordcloudGeneration} where the colors represent the associations of words to each cluster and the size of the words represents the scaled weight values of the prominent tokens ($w^*$). The unified wordcloud enables a quick understanding of the prominent tokens across all the clusters by accounting for their relative importance within each cluster as well as the number of samples that each cluster entails.
\subsection{Context-Aware Cluster Assignment}\label{sec:semisup_res}
\begin{figure}[!h]
    \centerline{\includegraphics[width=0.75\paperwidth]{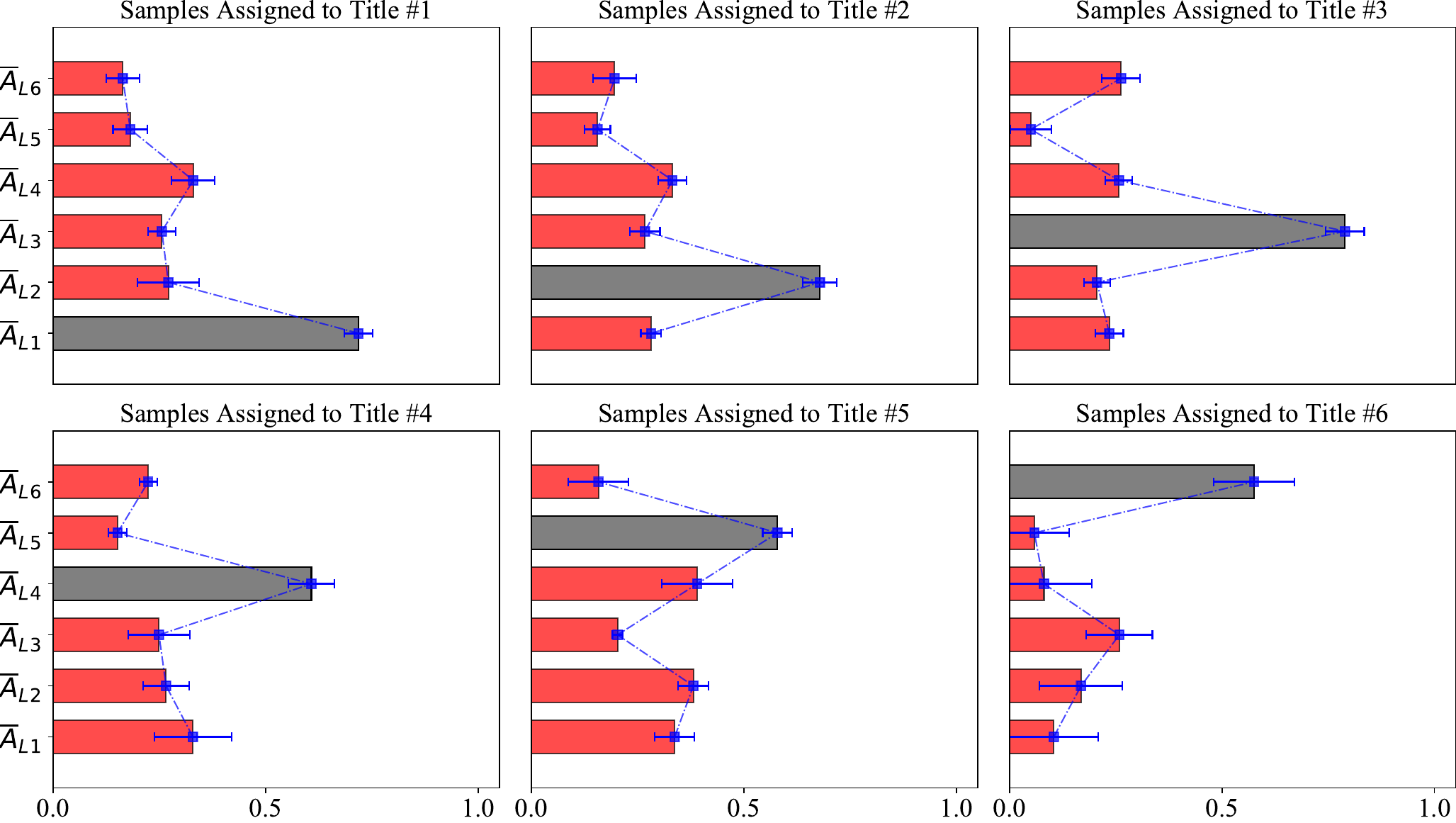}}
    \caption{The average (in red) as well as the standard deviation (in blue) values of the assignment matrix ($A$) for each input cluster label (title) across the samples assigned to each of the six input titles in table \eqref{tab:titles}.}
    \label{fig:cluster_assignment_hist2}
\end{figure}

\noindent In this section, we explain the way of using input titles in the context-aware cluster assignment task. Table \eqref{tab:titles} presents the list of a plausible set of input titles for the response categories in an open-response survey question for a chemistry class that the authors of this manuscript have manually written. In the survey, the teacher asks about one topic that the students would like to be reviewed again before their upcoming exam. The main difference between the cluster assignment task in this section and the clustering task in the previous section is that here the teacher has a certain set of bucket titles in mind where he/she wants to put the responses of students under those categories. As presented in subsection \eqref{sec:semi-supervised-clustering}, here in addition to the input responses we extract the embeddings of the the input titles in table \eqref{tab:titles} using the pre-trained language model as well. As described in subsection \eqref{sec:semi-supervised-clustering}, upon building the assignment matrix $A$ we assign to each input response the title with the highest value of cosine similarity in the embedding space.\\
\indent Figure \eqref{fig:cluster_assignment_hist2} shows the average (in red) as well as the standard deviation (in blue) values of the assignment matrix ($A$) for each cluster label (title) across the samples assigned to each of the six input titles in table \eqref{tab:titles}. It is worth noting that even though in figure \eqref{fig:cluster_assignment_hist2} we show the average values of the assignment matrix ($A$) for each input cluster label, the assignment happens at the input response and input title level where we assign to each input response the title with the highest value of cosine similarity in the embedding space.

\begin{table}[!h]
\centering
\arrayrulecolor{black}
\resizebox{0.6\columnwidth}{!}{
\begin{tabular}{|l|l|} 
\hline
\rowcolor{black} \textcolor{white}{ID} & \multicolumn{1}{c|}{\textcolor{white}{Input Titles}}                                                \\ 
\hline
\#1  & The chapter on molecular ionic and covalent bonds \\ 
\#2  & The chapter on atomic subparticles such as proton, electron, neutron           \\ 
\#3  & The chapter on thermodynamic concepts such as enthalpy and entropy           \\ 
\#4  & The chapter on acid and base reactions      \\ 
\#5  & The chapter on periodic table layout             \\ 
\#6  & The chapter on converting different units        \\ 
\arrayrulecolor{black}\hline
\end{tabular}
}
\caption{List of a plausible set of input titles for the response categories in an open-response survey question for a chemistry class that the authors of this manuscript have manually written.}
\label{tab:titles}
\end{table}
\section{Conclusion and Future Work}\label{sec:conclusion}
\noindent In this work, we presented a novel end-to-end context-aware framework to extract, aggregate, and annotate embedded semantic patterns in open-response survey data. Using a pre-trained natural language model, we encoded textual data at the sentence as well as word levels into semantic vectors. We then showed how the encoded vectors can get clustered either into an optimally tuned number of groups or into a set of groups with pre-specified titles. In the clustering task, we presented an approach to annotate the clusters in a context-aware manner. For both clustering as well as cluster assignment approaches, we presented the way of generating context-aware wordclouds that serve as a visual semantic representation of words taking into account their word-level and cluster-level importance. Finally, we provided some potential approaches for gathering descriptive statistics about the clusters in the clustering and cluster assignment tasks. It is worth noting that even though survey open-response analysis was the case study considered in this work, our proposed framework can be used for analyzing open-response survey data in any application. As one future direction, we intend to investigate extending our proposed framework to non-English languages using  multi-language pre-trained models and build a survey analysis tool that can work for different languages. As another future direction, we plan to evaluate our context-aware clustering framework on publicly available open-responses from students and consult with teachers to evaluate our generated insights. 

\appendix
\section{Appendix} \label{sec:appendix}
\renewcommand{\theequation}{\thesection.\arabic{equation}}
\setcounter{equation}{0}
\subsection{Silhouette Score} \label{sec:silhouette-score}
\noindent In this work, we use the silhouette score in order to find the optimal number of clusters in the clustering scenario. Silhouette score is a metric with values in the range of -1 and 1 and is used to evaluate a clustering outcome \cite{Pearce1977}, and basically measures the balance between cohesion (\ie the similarity of a sample to its own cluster) and separation (\ie the dissimilarity of a sample to other clusters) of the clustered samples. For an arbitrary sample $i$ in the dataset ($D$) that belongs to cluster $C_i$, we can define cohesion ($c$) and separation ($s$), respectively, as
\begin{subequations}
\begin{equation}
c(i) = \frac{1}{N_{C_i}-1}\sum_{\forall j \in C_i}d(i,j),\,\, \text{where}\,\,j\neq i\,,
\end{equation}    
\begin{equation}
s(i) = \min_{k \neq i} \Phi_k,\,\, \text{with}\,\, \Phi_k = \frac{1}{|C_k|}\sum_{\forall j \in C_k}d(i,j)\,,
\end{equation}
\end{subequations}
where $N_{C_i}$ is the number of samples within the cluster $C_i$ and $d(i,j)$ is the distance between points $i$ and $j$. $\Phi_k$ is the mean distance of sample $i$ in cluster $C_i$ to all the samples in cluster $C_k$, where separation $s(i)$ is then defined as the minimum of such mean distance across the clusters $C_k$ where $k\neq i$. For sample $i$ the lower the cohesion value $c(i)$, the more similar that sample is to its cluster, and the higher the separation value $s(i)$, the more dissimilar that sample is to all other clusters. For calculating the distance between two points $i$ and $j$, the function $d$ can be any distance metric such as Euclidean, Manhattan, or Chebyshev. In this work, we use the Euclidean distance function for calculating the cohesion and separation values. The silhouette value for sample $i$ is then defined as
\begin{equation}
  s(i)=\begin{cases}
    \frac{s(i)-c(i)}{\max[c(i),\,s(i)]} & N_{C_i}>1\\
    0 & N_{C_i}=1
  \end{cases}\,.
\end{equation}
Finally, the silhouette score for a given clustering configuration with $k$ number of clusters becomes the average of silhouette values of all the data samples and is defined as 
\begin{equation}
    SS^{(k)} = \frac{1}{N_D}\sum_{i\in D} s(i)\,,
\end{equation}
where $N_D$ is the number of data samples. 
\clearpage
\subsection{Pseudo-code for Context-Aware Clustering} \label{sec:Pseudocode1}
\noindent In this section, we present the pseudo-code for context-aware clustering presented in subsection \eqref{sec:un-supervised-clustering}. The function \textbf{getEmbeddingsFunc} takes as input sentences or words and returns their embedding vectors obtained from the pre-trained SBERT language model. The function \textbf{getClustersFunc} takes as its input the embedding vectors of the input responses and uses k-means algorithm and based on maximizing the silhouette score finds the optimum number of clusters and returns the clustered responses (as \textit{clustersDict}) as well as the centers of the clusters (as \textit{clustersCenters}). \textit{clustersDict} is a dictionary that assigns to each input response a corresponding cluster index. \textit{clustersDict} is a list with the length of number of clusters where each element of it is a vector corresponding to a cluster center with the vector dimension agreeing with the dimension of the embedding vectors, \ie $384$.
\begin{algorithm}[!h]
  \footnotesize
   \caption*{\textbf{Function 1a}: \footnotesize{Context-Aware Clustering}}
   \hrule 
    \begin{algorithmic}[1]
      \Function{\textbf{ClusteringFunc}}{inputResponses}
      \State Args: 
      \State ~~~~~$\triangleright$ inputResponses (list): list of input responses
      \State Returns: 
      \State ~~~~~$\triangleright$ clustersDict (dict): dictionary of \{cluster id:cluster samples\}
      \State inputResponsesEmbeddings = \textbf{getEmbeddingsFunc}(inputResponses)
      \State clustersDict, clustersCenters = \textbf{getClustersFunc}(inputResponsesEmbeddings)
      \State \Return clustersDict
       \EndFunction
\end{algorithmic}
\label{alg:SMOTE_binary1}
\end{algorithm}
\subsection{Pseudo-code for Context-Aware Cluster Assignment} \label{sec:Pseudocode3}
\noindent In this part, we present the pseudo-code for context-aware cluster assignment. The function \textbf{getEmbeddingsFunc} gets the embedding vectors for the input responses as well as the input titles, respectively, as \textit{inputResponsesEmbeddings} and \textit{clusterTitlesEmbeddings}. Next, the function \textbf{assignmentFunc} assigns to each input response the most cosine similar title in the embedding space as described in subsection \eqref{sec:semi-supervised-clustering}.
\begin{algorithm}[!h]
  \footnotesize
   \caption*{\textbf{Function 2}: \footnotesize{Context-Aware Cluster Assignment}}
   \hrule 
    \begin{algorithmic}[1]
      \Function{\textbf{ClusterAssignment}}{inputResponses, clusterTitles}
      \State Args: 
      \State ~~~~~$\triangleright$ inputResponses (list): list of input responses
      \State ~~~~~$\triangleright$ clusterTitles (list): list of cluster titles
      \State Returns: 
      \State ~~~~~$\triangleright$ clustersDict (dict): dictionary of \{cluster id:cluster samples\}
      \State inputResponsesEmbeddings = \textbf{getEmbeddingsFunc}(inputResponses)
      \State clusterTitlesEmbeddings = \textbf{getEmbeddingsFunc}(clusterTitles)
      \State clustersDict = \textbf{assignmentFunc}(inputResponsesEmbeddings, clusterTitlesEmbeddings)
      \State \Return clustersDict
       \EndFunction
\end{algorithmic}
\label{alg:SMOTE_binary1}
\end{algorithm}

\subsection{Pseudo-code for Context-Aware Cluster Labeling} \label{sec:Pseudocode2}
\noindent Here, we present the pseudo-code for context-aware cluster labeling (annotation). The function \textbf{processTokenizeFunc} gathers the tokens of all the samples in each cluster and pre-processes them. The \textbf{getEmbeddingsFunc} function extract the embedding vectors of the tokens. The \textbf{getProminentTokensFunc} returns the top \textit{numTopTokens}, \eg 5, prominent tokens of each cluster. As explained in subsection \eqref{sec:clusterAnnotation}, it evaluates the cosine similarity between the tokens and the cluster centers, and returns the top \textit{numTopTokens} with the largest cosine similarity values.
\begin{algorithm}[!h]
  \footnotesize
   \caption*{\textbf{Function 1b}: \footnotesize{Context-Aware Cluster Labeling}}
   \hrule 
    \begin{algorithmic}[1]
      \Function{\textbf{ClusterLabelingFunc}}{inputResponses, clustersDict, clustersCenters}
      \State Args: 
      \State ~~~~~$\triangleright$ inputResponses (list): list of input responses
      \State ~~~~~$\triangleright$ clustersDict (dict): dictionary of \{cluster id:cluster samples\}
      \State ~~~~~$\triangleright$ clustersCenters (list): list of clusters centers in the embedding space
      \State Returns: 
      \State ~~~~~$\triangleright$ clustersProminentTokens (list): list of prominent tokens of each cluster
      \State ~~~~~$\triangleright$ clustersProminentTokensWeights (list): list of weights of prominent tokens of each cluster
      \State clustersProminentTokens, clustersProminentTokensWeights = $[~]$, $[~]$
	  \For{clusterIdx in clustersDict.keys(\,)}
	  		\State clusterSamples            = clustersDict$[$clusterIdx$]$
	  		\State clusterCenter             = clustersCenters[clusterIdx]
	  		\State tokens = \textbf{processTokenizeFunc}(clusterSamples)
	  		\State tokensEmbeddings  = \textbf{getEmbeddingsFunc}(tokens)
	  		\State prominentTokens, weights = \textbf{getProminentTokensFunc}(clusterCenter, tokens, tokensEmbeddings, numTopTokens)
	  		\State clustersProminentTokens.append(prominentTokens)
	  		\State clustersProminentTokensWeights.append(weights)
	  \EndFor
      \State \Return (clustersProminentTokens, clustersProminentTokensWeights)
      \EndFunction
      \State 
      \Function{\textbf{getProminentTokensFunc}}{clusterCenter, tokens, tokensEmbeddings, numTopTokens}
      \State Args: 
      \State ~~~~~$\triangleright$ clusterCenter (vector): vector of cluster center in the embedding space
      \State ~~~~~$\triangleright$ tokens (list): list of tokens from the cluster samples
      \State ~~~~~$\triangleright$ tokensEmbeddings (list): list of tokens embeddings from the cluster samples
      \State ~~~~~$\triangleright$ numTopTokens (int): number of prominent tokens for each cluster
      \State Returns: 
      \State ~~~~~$\triangleright$ prominentTokens (list): a list of top `numTopTokens' tokens sorted w.r.t. weight in descending order
      \State ~~~~~$\triangleright$ weights (list): a list of top `numTopTokens' tokens weights sorted in descending order
      \State tokensWeights = $[~]$
	  \For{tokenEmbedding in tokensEmbeddings}
	  		\State tokensWeights.append(\textbf{CosineSim}(v1: tokenEmbedding, v2: clusterCenter)
	  \EndFor
	  \State sortedTokensWeights, sortedTokens = zip\,(\,$*$sorted\,(\,zip\,(\,tokensWeights, tokens\,)\,, descending\,=\,True)\,)
	  \State prominentTokens, weights = sortedTokens$[:\,$numTopTokens$]$,  sortedTokensWeights$[:\,$numTopTokens$]$
      \State \Return (prominentTokens, weights)
      \EndFunction
\end{algorithmic}
\label{alg:SMOTE_binary1}
\end{algorithm}

\small
\bibliographystyle{unsrt}
\bibliography{refs2} 

\end{document}